\documentclass[10pt,twocolumn,letterpaper]{article}

\usepackage{iccv}
\usepackage{times}
\usepackage{epsfig}
\usepackage{graphicx}
\usepackage{amsmath}
\usepackage{amssymb}
\usepackage{bm}
\usepackage{multirow}
\usepackage{mathbbol}
\usepackage{ulem}
\usepackage{subfigure}

\usepackage[pagebackref=true,breaklinks=true,colorlinks,bookmarks=false]{hyperref}
\usepackage[capitalize]{cleveref}
\crefname{section}{Sec.}{Secs.}
\Crefname{section}{Section}{Sections}
\Crefname{table}{Table}{Tables}
\crefname{table}{Tab.}{Tabs.}

\iccvfinalcopy 

\def\iccvPaperID{7269} 

\ificcvfinal\fi

\begin{document}

\title{LA-Net: Landmark-Aware Learning for Reliable Facial Expression Recognition under Label Noise}


\author{Zhiyu Wu, Jinshi Cui\thanks{Corresponding author} \\
School of Intelligence Science and Technology, Peking University \\
{\tt\small wuzhiyu@pku.edu.cn, cjs@cis.pku.edu.cn}
}

\maketitle
\ificcvfinal\thispagestyle{empty}\fi

\begin{abstract}
Facial expression recognition (FER) remains a challenging task due to the ambiguity of expressions. The derived noisy labels significantly harm the performance in real-world scenarios. To address this issue, we present a new FER model named Landmark-Aware Net~(LA-Net), which leverages facial landmarks to mitigate the impact of label noise from two perspectives. Firstly, LA-Net uses landmark information to suppress the uncertainty in expression space and constructs the label distribution of each sample by neighborhood aggregation, which in turn improves the quality of training supervision. Secondly, the model incorporates landmark information into expression representations using the devised expression-landmark contrastive loss. The enhanced expression feature extractor can be less susceptible to label noise. Our method can be integrated with any deep neural network for better training supervision without introducing extra inference costs. We conduct extensive experiments on both in-the-wild datasets and synthetic noisy datasets and demonstrate that LA-Net achieves state-of-the-art performance. 
\end{abstract}

\section{Introduction}
\label{sec:introduction}

Facial expression recognition (FER) is a fundamental task in the computer vision community, as it holds significance in various applications, such as human-computer interaction and fatigue driving detection. Given its wide applicability, researchers have developed multiple FER models in recent years~\cite{shan2009facial, kumari2015facial, li2020deep}. Despite the progress made so far, FER remains challenging due to noisy labels. Specifically, individuals may interpret one expression differently, resulting in inconsistent annotations. Besides, facial expressions have inherent inter-class similarity, which further exacerbates the problem of label noise. Accordingly, mitigating label noise has become one of the primary tasks in FER.

Mainstream noise-tolerant FER models can be divided into two groups: selecting clean samples and using label distributions as auxiliary training targets. For example, SCN~\cite{wang2020suppressing} estimates the uncertainty of each sample and corrects the mislabeled samples during training. Moreover, DMUE~\cite{she2021dive} utilizes multiple learnable branches to estimate the label distribution of each sample. While these methods promote FER performance under label noise, they still encounter some problems. Firstly, sample selection methods assume that the neural network will fit clean samples before overfitting noisy labels~\cite{arpit2017closer, zhang2021understanding}, which can lead to confusion between hard samples and noisy samples. Secondly, label distributions calculated based solely on expression information can still be noisy, since corrupt labels will disrupt both class and feature spaces~\cite{li2022neighborhood}.

To address the above problems, this paper proposes a new noise-tolerant FER model, called Landmark-Aware Net~(LA-Net). The overview of LA-Net is plotted in~\cref{fig:model}. LA-Net leverages facial landmark information to combat label noise based on the underlying assumption that expressions with analogous landmark patterns probably belong to the same emotion category. The model comprises two key modules: label distribution estimation~(LDE) and expression-landmark contrastive loss~(EL Loss).

LDE calculates the label distribution of each sample and uses it as an auxiliary supervision signal. Based on the assumption that expressions should have similar emotions to their neighbors in feature space, LDE identifies neighbors in both expression and landmark spaces for each sample. The landmark information is utilized to correct the errors in the expression space. The module then learns pairwise contribution scores and performs neighborhood aggregation to obtain target label distributions. Furthermore, to mitigate the impact of batch division on online aggregation, the target label distributions are summed over previous epochs using exponential moving average~(EMA).

EL Loss incorporates landmark information into expression representations to develop a robust backbone that is less susceptible to label noise. The algorithm treats landmarks and expressions as two views of facial images and establishes interactions between them via supervised contrastive learning~(SCL)~\cite{khosla2020supervised}. However, traditional SCL uses one-hot labels to select positive and negative pairs, thus performing poorly in the presence of label noise. Accordingly, our EL Loss designs a new pair selection strategy based on the label distributions to enable noise-tolerant SCL. Specifically, it first assigns pseudo-labels to confident images and takes the rest as unsupervised samples. The module then uses the expression and landmark features of the same images, or images with the identical pseudo-label, as positive pairs and all other combinations as negative pairs.

The proposed modules are solely used during training and thus incur no extra costs in deployment. Overall, our contributions can be summarized as follows:

(1) We present a landmark-aware FER model, named LA-Net, which leverages facial landmarks to alleviate the label noise issue.

(2) The LDE module uses landmark information to correct the errors in expression space and finds a set of neighbors to construct the label distribution of each sample.

(3) EL Loss devises noise-tolerant supervised contrastive learning and strengthens the expression feature extractor via expression-landmark interactions.

(4) LA-Net achieves state-of-the-art performance on both in-the-wild datasets and synthetic noisy datasets.

\section{Related Work}
\label{sec:related-work}

\subsection{Facial Expression Recognition}
Researchers have proposed many algorithms to improve FER performance~\cite{shan2009facial, kumari2015facial, li2020deep}. At the outset, handcrafted features including HOG~\cite{dalal2005histograms} and SIFT~\cite{ng2003sift} are applied to analyze expressions, whereas they perform poorly in the presence of strong illumination changes, large pose variations, and occlusions. Subsequently, learning-based methods advance the research mainly in two aspects: combating label noise and locating key areas. To address label ambiguity, SCN~\cite{wang2020suppressing} evaluates the uncertainty of each sample and corrects mislabeled training samples on-the-fly. DMUE~\cite{she2021dive}, on the other hand, applies multiple branches to calculate the label distribution of each sample. Regarding region-based models, RAN~\cite{wang2020region} uses self-attention to capture the importance of each facial area. TransFER~\cite{xue2021transfer} further introduces a dropout-like strategy to extract diverse key areas. 

Concurrent with our work, LDLVA~\cite{le2023uncertainty} also utilizes auxiliary facial information~(valance-arousal) to alleviate label noise. Nonetheless, they use the extra information in a plug-and-play manner, while we take it as an auxiliary task and leverage supervised contrastive learning adapted to the label noise scenario to develop a robust feature extractor. Moreover, we combine the landmark and expression information to construct label distributions for training images and mitigate the impact of batch division by exponential moving average~(EMA).

\subsection{Learning with Noisy Labels}
Label noise is a common issue in datasets, making learning with noisy labels a crucial topic in the community. Current research can be categorized into two groups: designing robust loss functions and selecting clean samples for training. Regarding the noise-tolerant loss functions, Zhang~\textit{et al.}~\cite{zhang2018generalized} devise generalized cross-entropy loss to suppress label noise. Moreover, Patrini~\textit{et al.}~\cite{patrini2017making} propose a loss correction approach for robust training. The sample selection strategy is based on the small-loss assumption that neural networks fit clean samples before overfitting noisy labels. Specifically, Han~\textit{et al.}~\cite{han2018co} develop a co-teaching method that selects low-loss samples in two models to filter errors caused by noisy labels. Malach~\textit{et al.}~\cite{malach2017decoupling} propose to train two models and perform updates only in case of disagreement between them, thereby keeping the effective number of noisy labels seen throughout the training process at a constant rate.

This paper aims to address the label noise in FER. The proposed LA-Net leverages facial landmarks to construct label distributions for training samples as well as strengthen the expression feature extractor via noise-tolerant supervised contrastive learning.

\subsection{Contrastive Learning}
Recently, Contrastive Learning (CL)~\cite{wu2018unsupervised, he2020momentum, chen2020improved, chen2020simple, wang2022rethinking, khosla2020supervised}, which is based on the Siamese network~\cite{bromley1993signature}, achieved great progress in unsupervised learning. Two prominent CL frameworks are SimCLR~\cite{chen2020simple} and MoCo~\cite{he2020momentum, chen2020improved}. SimCLR takes two random augmented views of the same image as positive pairs and different images as negative pairs, forming an instance discrimination task. Moreover, MoCo maintains a memory bank to increase negative samples and turns one branch of the Siamese network into a momentum encoder to improve the consistency of the memory bank. In addition to the self-supervised scenario, Khosla~\textit{et al.}~\cite{khosla2020supervised} utilize the contrastive loss in the supervised setting~(SCL) and achieve leading performance on multiple benchmarks.

Our EL Loss draws inspiration from supervised contrastive learning and implements interactions between landmarks and expressions to develop a more robust feature extractor. Moreover, we utilize label distribution as an alternative criterion for pair selection, which makes our method more resistant to label noise.

\begin{figure*}
    \centering
    \includegraphics[width=\textwidth]{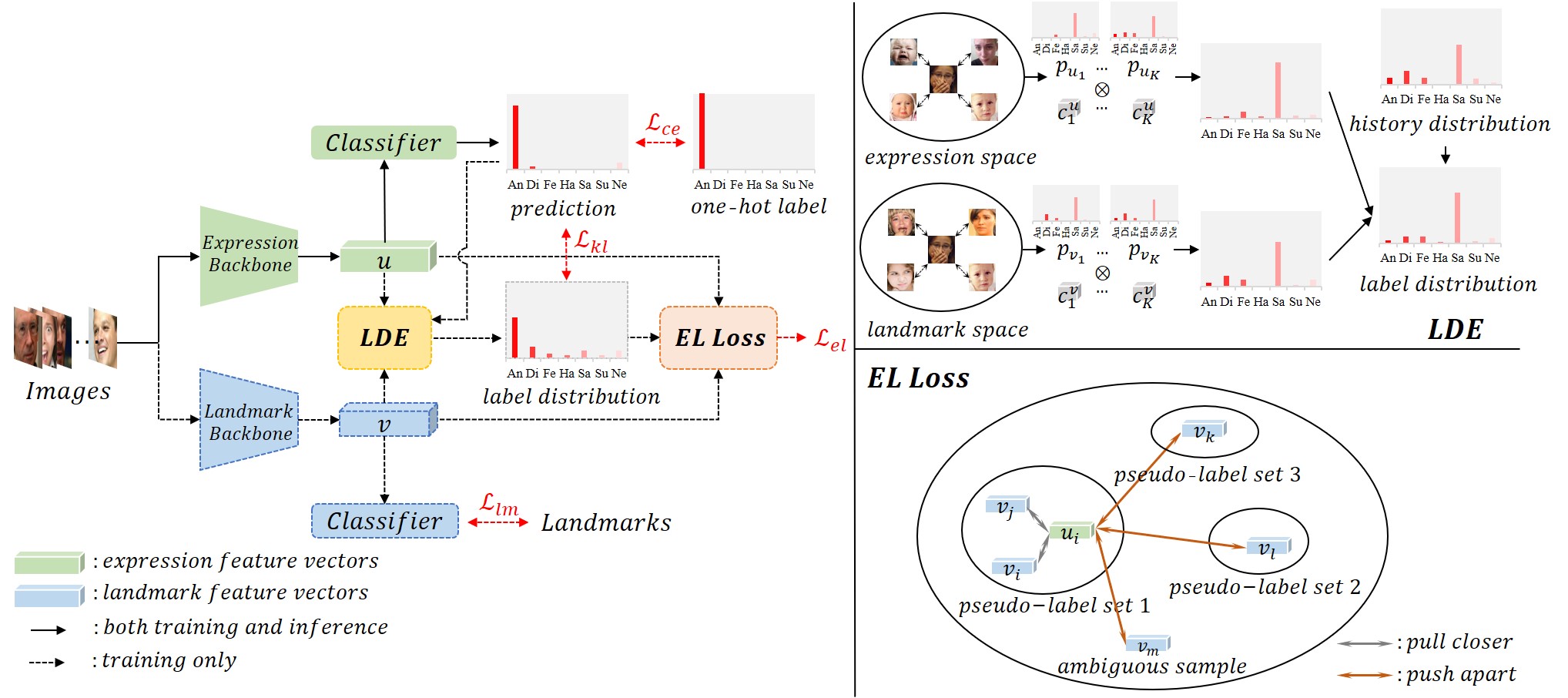}
    \caption{We present the pipeline of LA-Net on the left, with dashed lines indicating components used only during training and solid lines indicating those used in both training and inference. We zoom in on the structure of label distribution estimation~(LDE) and expression-landmark contrastive loss~(EL Loss) at the upper right and lower right of the figure, respectively. Regarding EL Loss, we provide an example of selecting positive and negative pairs for the expression feature of sample $\bm{x_i}$.}
    \label{fig:model}
\end{figure*}

\section{Methodology}
\label{sec:methodology}
We first introduce the notations that will be used in this paper. Let $\bm{x}$ be the instance variable and $\bm{x_i}$ be the $i$-th sample. We denote the class set as $\mathcal{Y}=\{y_1, ..., y_C\}$, where $C$ is the number of classes. Let $\bm{l_i}=(l_i^{y_1}, ..., l_i^{y_C})$ with $l_i^{y_j} \in \{0, 1\}$ and $||\bm{l_i}||_1=1$ indicate the one-hot label of sample $\bm{x_i}$. The label distribution of sample $\bm{x_i}$ is denoted as $\bm{d_i}=(d_i^{y_1}, ..., d_i^{y_C})$, where $d_i^{y_j} \in [0, 1]$ and $||\bm{d_i}||_1=1$. Let $\bm{u_i}$ and $\bm{v_i}$ be the expression and landmark features of sample $\bm{x_i}$, respectively. The prediction of sample $\bm{x_i}$ for the FER task is denoted as $\bm{p_i}=(p_i^{y_1}, ..., p_i^{y_C})$, where $p_i^{y_j} \in [0, 1]$ and $||\bm{p_i}||_1=1$.

The overview of LA-Net is shown in~\cref{fig:model}. The model consists of three main parts: backbone, landmark distribution estimation~(LDE), and expression-landmark contrastive loss~(EL Loss). Specifically, LA-Net first uses two backbones to extract the expression and landmark features respectively. The landmark localization part employs a fully connected layer as the classifier and minimizes the mean square error, denoted as $\mathcal{L}_{lm}$, during training. LDE identifies $2K$ neighbors ($K$ in expression space and $K$ in landmark space) for each sample and performs neighborhood aggregation to generate target label distributions, which in turn improve the quality of training supervision. Besides, EL Loss considers the similarity between landmarks and expressions and incorporates landmark information into expression representations using noise-tolerant supervised contrastive learning. In the following sections, we will describe the LDE module and EL Loss in detail.

\subsection{Label Distribution Estimation}
\label{sec:lde}
Given the presence of noisy labels, label distribution is a better descriptor for facial expressions than the one-hot label. To construct the target label distribution of each sample, previous works hold that expressions should have similar emotions to their neighbors in the feature space or a supporting space~\cite{le2023uncertainty, chen2020label}. Nonetheless, extreme label noise will disrupt both class and feature spaces. Accordingly, the LDE module uses landmark information to correct the errors in the expression space based on the assumption that images with similar landmark patterns should be assigned to the same emotion class.

Given the mini-batch $\mathcal{D}_{batch}=\{(\bm{x_i}, \bm{l_i})|i=1,2...n\}$, LDE calculates the label distributions in four steps. Firstly, for image $\bm{x_i}$, LDE adopts the $K$-Nearest Neighbor algorithm to identify its neighbors in the expression space, denoted as $N^u(i)$, based on the cosine similarity. 
\begin{align}
    s_{i, j}^u &= \frac{\bm{u_i}\bm{u_j}^T}{||\bm{u_i}||||\bm{u_j}||}\\
    N^u(i) &= KNN(\mathcal{D}_{batch};K;s_i^u) 
\end{align}
where $s_{i, j}^u$ denotes the similarity between $\bm{x_i}$ and $\bm{x_j}$ in the expression space. The neighbor set in the landmark space, denoted as $N^v(i)$, can be generated in an identical way. Then, LDE evaluates the pairwise contribution scores in the expression space as follows:
\begin{align}
    c_{i, j}^u = \text{Sigmoid}(f([g_1(\bm{u_i};\theta_1), g_2(\bm{u_j};\theta_2)];\theta))
\end{align}
where $c_{i,j}^u$ denotes the contribution of $\bm{x_j}$ to $\bm{x_i}$ in the expression space, $[]$ denotes concatenation, $f$, $g_1$, and $g_2$ are three MLPs with learnable parameters $\theta$, $\theta_1$, and $\theta_2$. The contribution scores in the landmark space, denoted as $c_{i, j}^v$, can be obtained in the same way. Based on the scores, the module performs neighborhood aggregation in each space and produces the target label distributions by mean pooling.
\begin{align}
    \bm{d_i} = \frac{1}{2}(\frac{\sum_{k \in N^u(i)} c^u_{i, k} \ \bm{p_k}}{\sum_{k \in N^u(i)} c^u_{i, k}} + \frac{\sum_{k \in N^v(i)} c^v_{i, k} \ \bm{p_k}}{\sum_{k \in N^v(i)} c^v_{i, k}})
\end{align}
As a result, the label distributions incorporate both expression and landmark information and thus are less susceptible to noisy labels. However, the above online aggregation approach can be affected by batch division, resulting in noisy and erratic label distributions in some cases. In particular, one batch may contain an excessive number of corrupt labels, causing the neighborhood of each sample to be extremely noisy. To address this issue, LDE sums up the targets over previous epochs using exponential moving average~(EMA). This allows us to compute the target label distribution of image $\bm{x_i}$ in the $e$-th epoch as follows:
\begin{align}
    d_i^{[e]} = \omega \ d_i^{[e-1]} + (1 - \omega) \ \bm{d_i}
\end{align}
where $\omega$ denotes the decay of previous targets. LA-Net leverages both one-hot labels and label distributions as supervision and minimizes the following loss functions:
\begin{align}
    \mathcal{L}_{ce} &= -\frac{1}{n} \sum_{i=1}^n \sum_{j=1}^C l_i^{y_j} \text{log}(p_i^{y_j})\\
    \mathcal{L}_{kl} &= -\frac{1}{n} \sum_{i=1}^n \sum_{j=1}^C d_i^{[e], y_j} \text{log} \frac{d_i^{[e],y_j}}{p_i^{y_j}}
\end{align}

\subsection{Expression-Landmark Contrastive Loss}
\label{sec:el-loss}
The LDE module estimates label distributions for training samples to enhance resistance to corrupt labels. Besides, a knowledgeable feature extractor can also help mitigate the label noise. As such, LA-Net leverages facial landmarks to strengthen the expression feature extractor. Specifically, we create expression-landmark pairs and utilize supervised contrastive learning~(SCL)~\cite{khosla2020supervised}, denoted as expression-landmark contrastive loss~(EL Loss), to implement interactions between these two facial modalities. However, traditional SCL constructs positive and negative pairs based on the one-hot labels in datasets, thus performing poorly in the presence of label noise. To address this problem, EL Loss uses label distributions as an alternative criterion for pair selection, enabling noise-tolerant supervised contrastive learning. We describe the pair selection process for an anchor expression feature vector in~\cref{fig:model}, and provide further details below.

Given sample $\bm{x_i}$, EL Loss first derives its pseudo-label $\hat{l_i}$ from the target label distribution in the current epoch:
\begin{align}
    \hat{l_i} &= 
    \begin{cases}
        argmax(d_i^{[e]}) & \text{if } max(d_i^{[e]} > \delta) \\
        -1 & \text{otherwise}
    \end{cases} 
\end{align}
where $\delta$ denotes the confidence threshold for separating confident and ambiguous samples. We then treat the expression and landmark features of confident samples with the same pseudo-label as positive pairs. Moreover, to suppress the uncertainty of ambiguous samples, only expression and landmark features of the same ambiguous image are utilized as positive pairs. In other words, the indication set $\mathcal{I}(i)$ for the positive pairs of sample $\bm{x_i}$ is computed by:
\begin{align}
    \mathcal{I}(i) &= 
    \begin{cases}
        \{j \ |\ \hat{l_j}=\hat{l_i}\} & \text{if } \hat{l_i}\neq -1 \\
        \{i\} & \text{otherwise}
    \end{cases}
\end{align}
In addition, we take the remaining sample combinations as negative pairs and denote them as $\Bar{\mathcal{I}}$. Following traditional contrastive learning~\cite{he2020momentum, chen2020improved, chen2020simple}, the module then performs non-linear projection to obtain query expression features.
\begin{align}
    q_i^u = g(\bm{u_i};\theta)
\end{align}
where $g$ is a two-layer perceptron with learnable parameters $\theta$. Query landmark features $q_i^v$, key expression features $k_i^u$, and key landmark features $k_i^v$ are calculated similarly. Finally, EL Loss implements interactions between expressions and landmarks by pulling positive pairs closer and pushing negative pairs farther apart.
{\small
\begin{align}
    \mathcal{L}_1 &= \frac{1}{n} \sum_{i=1}^n \frac{-1}{|\mathcal{I}(i)|} \sum_{j \in \mathcal{I}(i)} \text{log} \frac{\text{exp}( q_i^u \bm{\cdot} k_j^v / \tau)}{\sum\limits_{m \in \Bar{\mathcal{I}}(i) \cup \{j\}} \text{exp}(q_i^u \bm{\cdot} k_m^v / \tau)} \\
    \mathcal{L}_2 &= \frac{1}{n} \sum_{i=1}^n \frac{-1}{|\mathcal{I}(i)|} \sum_{j \in \mathcal{I}(i)} \text{log} \frac{\text{exp}(q_i^v \bm{\cdot} k_j^u / \tau)}{\sum\limits_{m \in \Bar{\mathcal{I}}(i) \cup \{j\}} \text{exp}(q_i^v \bm{\cdot} k_m^u / \tau)} \\
    \mathcal{L}_{el} &= \mathcal{L}_1 + \mathcal{L}_2
\end{align}}
where $a \bm{\cdot} b$ denotes the cosine similarity between $a$ and $b$, $\tau$ indicates a temperature parameter. Moreover, we follow the MoCo framework~\cite{he2020momentum, chen2020improved} and use memory banks to increase negative pairs.

\subsection{Overall Loss}
LA-Net minimizes the following loss function during training:
\begin{align}
    \mathcal{L} = \mathcal{L}_{ce} + \mathcal{L}_{lm} + \alpha \mathcal{L}_{kl} + \beta \mathcal{L}_{el}
\end{align}
where $\alpha$ indicates the importance of label distributions and $\beta$ denotes the contribution of EL Loss.

\section{Experiments}
\label{sec:experiments}

\subsection{Datasets}
RAF-DB~\cite{li2017reliable} consists of 30,000 images with basic or compound labels. Following previous works, we only use the images annotated with six basic expressions (\textit{anger}, \textit{disgust}, \textit{fear}, \textit{happiness}, \textit{sadness}, \textit{surprise}) and \textit{neutral}, where 12,271 are used for training and 3,068 for testing. 

FERPlus~\cite{barsoum2016training} is an improved version of FER2013~\cite{goodfellow2013challenges}. It contains 28,709/3,589/3,589 training/validation/testing images. Each image is labeled by ten experts and assigned to one of eight classes~(six basic expressions, \textit{neutral}, and \textit{contempt}). We use the class with the most votes as the label. 

AffectNet~\cite{mollahosseini2017affectnet} is the largest and most challenging FER dataset. It contains more than 280K training images and 4,000 testing images, which are classified into the same eight classes as FERPlus. Previous works vary in the use of AffectNet (with or without \textit{contempt}), and we utilize 8-class AffectNet by default in the experiments.

Following previous works, we report the overall accuracy of the testing set for all datasets. 

\subsection{Implementation Details}
We perform experiments on 4 NVIDIA RTX 3080Ti GPUs and use ResNet-18~\cite{he2016deep} pretrained on MS-Celeb-1M~\cite{guo2016ms} as the backbone. Regarding data processing, we resize all images to $224 \times 224$ pixels and use HRNet~\cite{sun2019deep, wang2020deep} to generate ground truth landmarks. To alleviate class imbalance, we adopt progressively balanced sampling~\cite{kang2019decoupling}. Besides, on-the-fly data augmentation techniques, including random cropping, random horizontal flipping, random erasing, and random color jitter, are employed to enhance the generalization performance of LA-Net. During training, we set the batch size to 128 and use the Adam optimizer~\cite{kingma2014adam} with an initial learning rate of 1e-3. For all datasets, we train the model for 80 epochs and linearly decrease the learning rate to 0. Moreover, we conduct grid searches to determine all parameters. Specifically, we set the number of neighbors $K$ to 8, the decay of targets $\omega$ to 0.9, the temperature $\tau$ to 0.1, and the confidence threshold $\delta$ to 0.7. Besides, weights $\alpha$ and $\beta$ in the overall objective are set to 1.0 and 0.1.

\begin{table}[t]
    \footnotesize
    \centering
    \begin{tabular}{c|c|c c c}
    \hline
    Noise & Method & RAF-DB & FERPlus & AffectNet \\
    \hline
    \hline
    \multirow{5}*{10\%} & Baseline & 81.01 & 83.29 & 57.24\\
    ~ & SCN~\cite{wang2020suppressing} & 82.18 & 84.28 & 58.58 \\
    ~ & RUL~\cite{zhang2021relative} & 86.17 & 86.93 & 60.54  \\
    ~ & EAC~\cite{zhang2022learn} & 88.02 & 87.03 & 61.11 \\
    ~ & LA-Net~(Ours) & \textbf{88.75±0.11} & \textbf{88.02±0.08} & \textbf{62.85±0.13}\\
    \hline
    \multirow{5}*{20\%} & Baseline & 77.98 & 82.34 & 55.89\\
    ~ & SCN~\cite{wang2020suppressing} & 80.10 & 83.17 & 57.25\\
    ~ & RUL~\cite{zhang2021relative} &  84.32 & 85.05 & 59.01 \\
    ~ & EAC~\cite{zhang2022learn} & 86.05 & 86.07 & 60.29\\
    ~ & LA-Net~(Ours) & \textbf{87.12±0.16} & \textbf{86.85±0.14} & \textbf{61.72±0.21}\\
    \hline
    \multirow{5}*{30\%} & Baseline &  75.50 & 79.77 & 52.16 \\
    ~ & SCN~\cite{wang2020suppressing} & 77.46 & 82.47& 55.05 \\
    ~ & RUL~\cite{zhang2021relative} & 82.06 & 83.90 & 56.93 \\
    ~ & EAC~\cite{zhang2022learn} & 84.42 & 85.44 & 58.91\\
    ~ & LA-Net~(Ours) & \textbf{85.33±0.18} & \textbf{86.01±0.19} & \textbf{60.82±0.21}\\
    \hline
    \end{tabular}
    \caption{Evaluation~(\%) on synthetic noisy datasets.}
    \label{tab:synthetic-noise}
\end{table}

\subsection{Experiments on Synthetic Noisy Datasets}
Noisy labels, caused by ambiguous facial expressions, significantly harm FER performance in real-world scenarios. To address this issue, LA-Net leverages landmark information to estimate label distributions for training samples and strengthen the expression feature extractor. We compare LA-Net with previous noise-tolerant FER models on three synthetic noisy datasets with varying levels of noise (10\%, 20\%, and 30\%). To inject synthetic noise, we follow prior studies~\cite{wang2020suppressing, zhang2021relative, zhang2022learn} and randomly flip the one-hot label to other categories. Additionally, we repeat the experiments five times due to the randomness of label noise injection and report the mean and standard deviation of overall accuracy.

\cref{tab:synthetic-noise} shows that LA-Net consistently outperforms other noise-tolerant FER algorithms in all settings. Compared to the current state-of-the-art model EAC~\cite{zhang2022learn}, LA-Net achieves an average improvement of 0.91\%, 0.82\%, and 1.70\% on RAF-DB, FERPlus, and AffectNet, respectively. Additionally, LA-Net becomes more preferred as the noise ratio increases. Specifically, compared to the baseline, the model promotes the performance by 5.61\% and 8.66\% on AffectNet with 10\% and 30\% noise, respectively. In conclusion, the experimental results in \cref{tab:synthetic-noise} demonstrate that the proposed approach is effective in mitigating label noise.

In addition to the aforementioned symmetric noise, we further test the more challenging asymmetric noise, where the label is flipped to its most similar class based on the confusion matrix. As shown in \textit{Appendix A}, LA-Net consistently outperforms previous models when dealing with asymmetric noise, indicating its effectiveness.

\begin{table}[t]
    \small
    \centering
    \begin{tabular}{c c c c|c c}
    \hline
    \multirow{2}*{LD} & \multirow{2}*{LM} & \multirow{2}*{EL} & \multirow{2}*{PL} & RAF-DB & RAF-DB \\
    ~ & ~ & ~ & ~ & (original) & (30\% noise) \\
    \hline
    \hline
    - & - & - & - & 87.06 & 75.50 \\
    \hline
    \checkmark & - & - & - & 88.43 & 79.17 \\
    \checkmark & \checkmark & - & - & 89.04 & 82.37 \\
    \hline
    \checkmark & \checkmark & \checkmark & - &  90.48 & 83.21 \\
    \checkmark & \checkmark & \checkmark & \checkmark & 90.81 & 85.33 \\
    \hline
    \end{tabular}
    \caption{Component analysis~(\%). (LD: label distribution generated using only expression information; LM: landmark information in LDE; EL: expression-landmark contrastive loss; PL: pseudo-labels in EL Loss.)}
    \label{tab:ablation}
\end{table}

\begin{table*}[t]
    \small
    \centering
    \begin{tabular}{c|c c c c}
    \hline
    Method & RAF-DB & AffectNet (7 classes) & AffectNet (8 classes) & FERPlus \\
    \hline
    \hline
    RAN~\cite{wang2020region} & 86.90 & - & 59.50 & 89.16 \\
    SCN~\cite{wang2020suppressing} & 87.03 & - & 60.23 & 89.39 \\
    DACL~\cite{farzaneh2021facial} & 87.78 & 65.20 & - & - \\
    KTN~\cite{li2021adaptively} & 88.07 & 63.97 & - & 90.49 \\
    FDRL~\cite{ruan2021feature} & 89.47 & - & - & - \\
    ARM~\cite{shi2021learning} & 90.42 & 65.20 & 61.33 & - \\
    DMUE~\cite{she2021dive} & 88.76 & - & 62.84 & 88.64 \\
    EAC~\cite{zhang2022learn} & 89.99 & 65.32 & - & 89.64 \\
    LDLVA~\cite{le2023uncertainty} & 90.51 & 66.23 & - & - \\
    TransFER~\cite{xue2021transfer} & 90.91 & 66.23 & - & 90.83\\
    \hline
    LA-Net (ResNet-18) & 90.81 & 67.09 & 64.24 & 91.39 \\
    LA-Net (ResNet-50) & \textbf{91.56} & \textbf{67.60} & \textbf{64.54} & \textbf{91.78}\\
    \hline
    \end{tabular}
    \caption{Comparison (\%) with state-of-the-art methods on original datasets.}
    \label{tab:sota}
\end{table*}

\subsection{Ablation Study}
\label{sec:grid_search}
\noindent\textbf{Contribution of each component.}
To quantify the contribution of the proposed modules, we conduct an ablation study on both original and noisy RAF-DB. For simplicity, we refer to the performance on these two datasets as $(a, b)$ in the following. As shown in \cref{tab:ablation}, complete LDE~(line 3) promotes performance by (1.98\%, 6.87\%) compared to the baseline in line 1. EL Loss~(line 5), on the other hand, yields gains of (1.77\%, 2.96\%) in contrast to the model in line 3. These two modules leverage landmark information from different perspectives and mitigate real-world and synthetic noise effectively. Besides, comparing line~2 and line~3, landmark information in LDE brings the improvement of (0.61\%, 3.20\%), indicating its importance, especially in the presence of severe noise. Additionally, the pseudo-labels~(line 5) enable noise-tolerant supervised contrastive learning and achieve an improvement of 2.12\% on corrupt RAF-DB compared to the EL Loss using one-hot labels~(line 4). Overall, the results in \cref{tab:ablation} demonstrate the effectiveness of the proposed modules and the advantages of their combination in LA-Net.

\begin{figure}[t]
    \centering
    \includegraphics[width=0.4\textwidth]{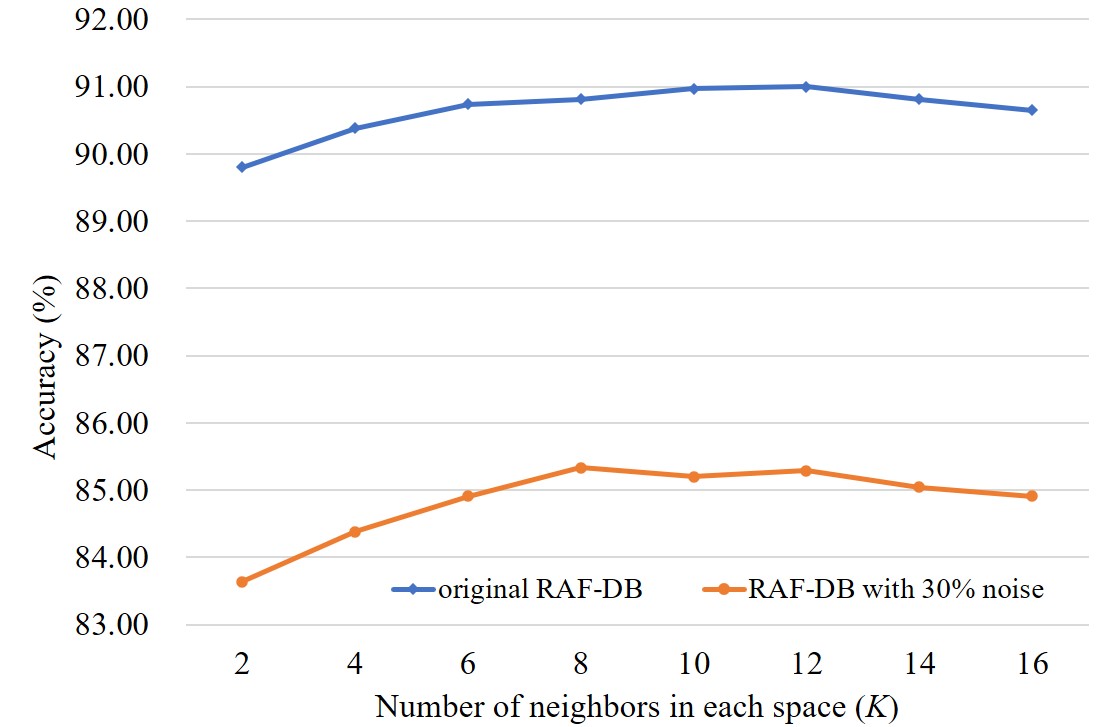}
    \caption{Performance on original and noisy RAF-DB with different number of neighbors in each space~($K$).}
    \label{fig:neighbor-num-K}
\end{figure}

\noindent\textbf{Number of nearest neighbors in each space $K$.} We explore the effect of $K$ on the performance in~\cref{fig:neighbor-num-K}. For original RAF-DB, LA-Net achieves optimal performance with $K$=12. Smaller or larger $K$ can lead to slight degradation, as the former may result in inaccurate target label distributions, while the latter can cause excessive noisy neighbors. The choice of $K$ has a similar effect on both original and noisy RAF-DB. The only exception is that LA-Net achieves the best performance on the noisy RAF-DB with $K$=8 since it contains more corrupt labels. Therefore, we set $K$ to 8 throughout experiments to alleviate label noise.

\subsection{Experiments on Original Datasets}
While the original datasets are considered "clean", they inevitably contain noisy labels due to the ambiguity of expressions. Hence, we compare LA-Net with state-of-the-art methods on original datasets to verify its resistance to real-world uncertainty. Moreover, we consider both 7-class and 8-class AffectNet for a fair comparison.

We compare the performance of LA-Net with several state-of-the-art FER models in~\cref{tab:sota}. Among the existing models, TransFER~\cite{xue2021transfer} utilizes the powerful Vision Transformer~\cite{dosovitskiy2020image} structure to improve performance, while the others enhance FER performance by locating key areas or alleviating label noise. Compared to existing models, LA-Net gains improvements of 0.65\%, 1.37\%, 1.70\%, and 0.95\% on RAF-DB, 7-class AffectNet, 8-class AffectNet, and FERPlus, respectively. Moreover, the proposed model also exhibits substantial advantages in challenging cases, such as 7-class and 8-class AffectNet. Our findings indicate that LA-Net is a promising approach for handling ambiguous expressions in real-world scenarios.

\begin{table}[t]
    \small
    \centering
    \begin{tabular}{c|c c}
    \hline
    Method & RAF-DB & AffectNet~(7 classes) \\
    \hline
    \hline
    AIR~\cite{azadi2015auxiliary} & 67.37 & 54.23 \\
    NAL~\cite{goldberger2017training} & 84.22 & 55.97 \\
    IPA2LT~\cite{zeng2018facial} & 83.80 & 57.85 \\
    LDL-ALSG~\cite{chen2020label} & 85.53 & 59.35 \\
    LDLVA~\cite{le2023uncertainty} & 87.26 & 62.89 \\
    \hline
    LA-Net (Ours) & \textbf{88.10} & \textbf{65.43} \\
    \hline
    \end{tabular}
    \caption{Experiments with inconsistent labels~(\%). Following previous works~\cite{zeng2018facial, chen2020label, le2023uncertainty}, we use the 7-class AffectNet to keep its label set consistent with RAF-DB.}
    \label{tab:inconsistent-labels}
\end{table}

\begin{figure*}[t]
    \centering
    \subfigure[Baseline model on RAF-DB with 30\% noise]{
        \includegraphics[width=0.3\textwidth]{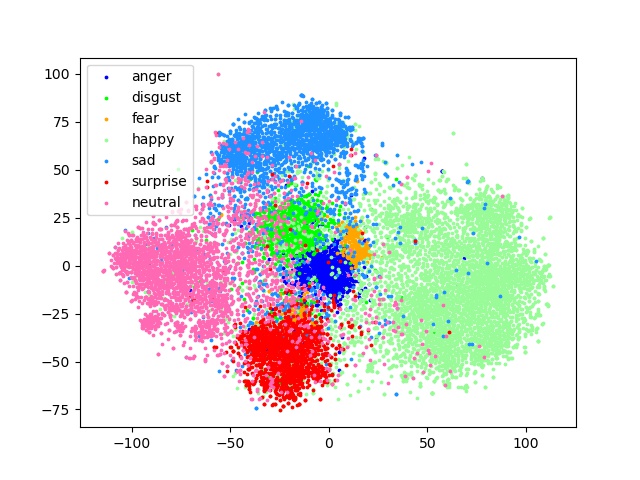}
        \label{raf-base-noise}
    }
    \subfigure[SCN~\cite{wang2020suppressing} on RAF-DB with 30\% noise]{
        \includegraphics[width=0.3\textwidth]{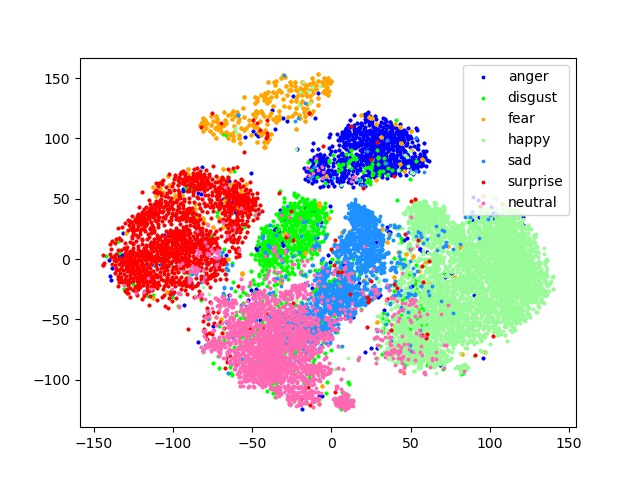}
        \label{scn-noise}
    }
    \subfigure[LA-Net on RAF-DB with 30\% noise]{
        \includegraphics[width=0.3\textwidth]{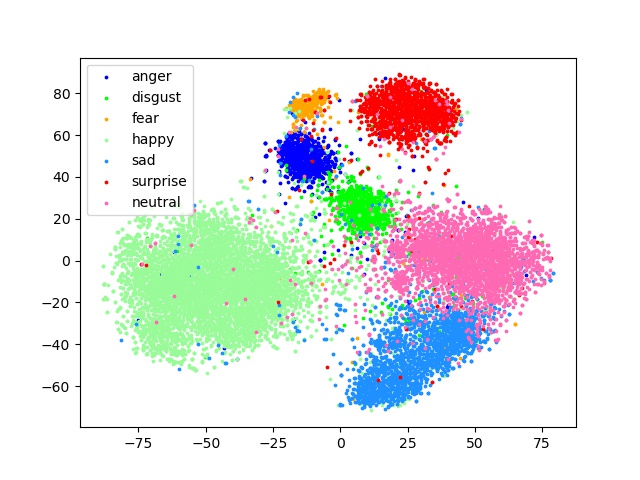}
        \label{raf-noise}
    }
    \caption{Feature visualization on noisy RAF-DB. The baseline model memorizes most of the noisy labels, resulting in indistinguishable feature clusters. SCN~\cite{wang2020suppressing} suppresses the noise to some extent, whereas it still overfits many corrupt labels. LA-Net pushes the noisy samples to the decision boundary and maintains clean clusters.}
    \label{fig:tsne}
\end{figure*}

\begin{figure*}[t]
    \centering
    \subfigure[Baseline model on RAF-DB with 30\% noise]{
        \includegraphics[width=0.3\textwidth]{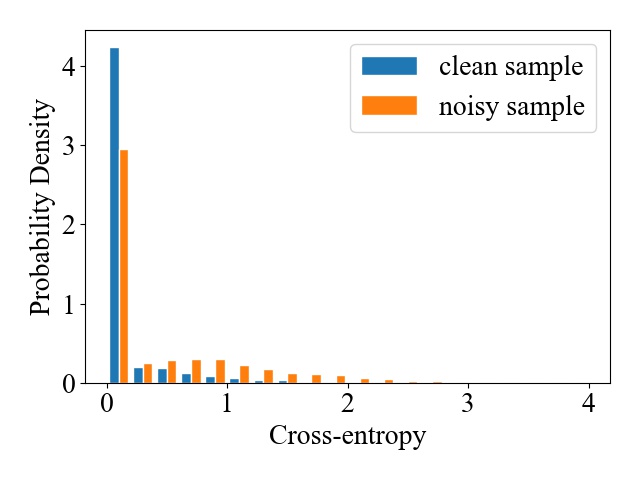}
        \label{baseline-loss}
    }
    \subfigure[SCN~\cite{wang2020suppressing} on RAF-DB with 30\% noise]{
        \includegraphics[width=0.3\textwidth]{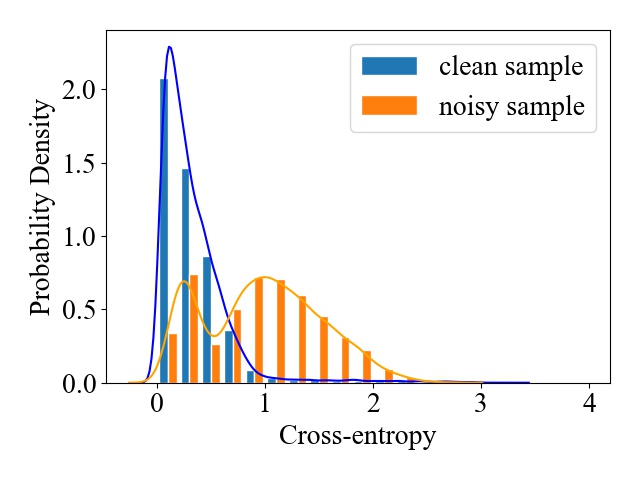}
        \label{scn}
    }
    \subfigure[LA-Net on RAF-DB with 30\% noise]{
        \includegraphics[width=0.3\textwidth]{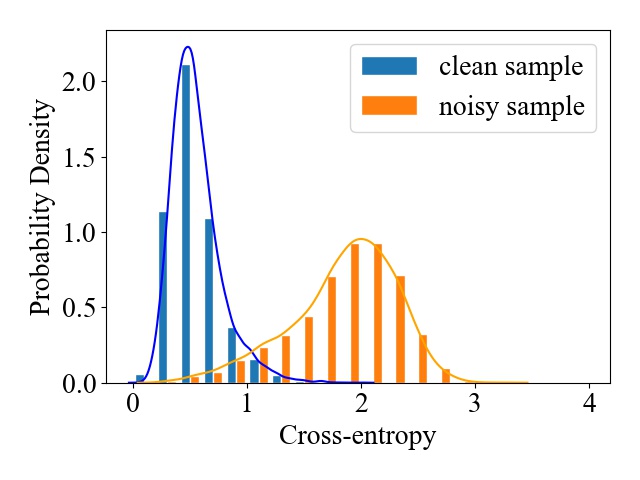}
        \label{lanet-loss}
    }
    \caption{Cross-entropy between predictions and one-hot labels after training.}
    \label{fig:loss-value}
\end{figure*}

\subsection{Experiments with Inconsistent Labels}
The performance of deep neural networks generally improves as more training samples are provided, but FER presents a unique challenge. Due to the subjective nature of interpreting facial expressions, individuals may assign different labels to the same expression, leading to inconsistent labels within a dataset and across different datasets. To evaluate the effectiveness of LA-Net in addressing this problem, we conduct a cross-dataset experiment following previous works~\cite{zeng2018facial, chen2020label, le2023uncertainty}. Specifically, we train our model on a mixed dataset composed of the training samples from RAF-DB and AffectNet and evaluate it on the testing set of both datasets. Note that we use the 7-class AffectNet to ensure the identical class set between two datasets.

We will analyze the results in~\cref{tab:inconsistent-labels} from two perspectives. Firstly, compared to LDLVA~\cite{le2023uncertainty}, which is the best available model for addressing label inconsistency, LA-Net achieves the improvement of 0.84\% and 2.54\% on RAF-DB and AffectNet, respectively. Secondly, LA-Net helps alleviate the performance drop caused by cross-dataset inconsistency. Specifically, comparing~\cref{tab:inconsistent-labels} and~\cref{tab:sota}, LA-Net encounters the degradation of 2.71\% on RAF-DB and 1.66\% on AffectNet, whereas LDLVA suffers from the drop of 3.25\% and 3.34\% on these two datasets, respectively. In summary, the improvement in two aspects demonstrates the effectiveness of LA-Net in dealing with inconsistent labels. Nonetheless, cross-dataset inconsistency remains a challenge for FER models and warrants further research.

\subsection{Visualization Analysis}
\noindent\textbf{High-dimensional features.} For an intuitive understanding of LA-Net, we use t-SNE~\cite{van2008visualizing} to plot the trained features of different models, including a baseline model, SCN~\cite{wang2020suppressing}, and our LA-Net, on RAF-DB with 30\% noise. As shown in~\cref{raf-base-noise}, the baseline model memorizes the noisy labels, resulting in adjacent feature clusters. SCN learns the uncertainty of each sample and proposes a relabeling mechanism to address label noise. Nonetheless, as depicted in~\cref{scn-noise}, it still overfits many noisy samples, such as the \textit{disgust} expressions in the cluster of \textit{anger}. In comparison, LA-Net, plotted in~\cref{raf-noise}, pushes noisy samples to the decision boundary and generates clean clusters. Moreover, our model achieves improved separation across categories and better compactness within each class. Overall, LA-Net prevents overfitting noisy samples and forms easy-to-distinguish category clusters in the presence of severe noise, indicating its effectiveness in dealing with noisy labels.

\noindent\textbf{Cross-entropy between predictions and one-hot labels.} We evaluate the ability of LA-Net to handle noisy labels by calculating the cross-entropy between predictions and one-hot labels of the training samples after training. As shown in~\cref{baseline-loss}, the baseline model memorizes almost all noisy labels, resulting in poor generalization ability. SCN~\cite{wang2020suppressing} alleviates the noise to some extent by relabeling, whereas it still overfits some corrupt labels since these noisy samples are not identified or mistakenly relabeled. In contrast, LA-Net, which leverages landmark information to construct label distributions and enhance expression feature extractor, could easily distinguish between clean and noisy samples after training. Overall, the results in~\cref{fig:loss-value} indicate that our model can suppress label noise to a large extent.

\begin{figure*}[t]
    \centering
    \subfigure{
        \includegraphics[width=0.3\textwidth]{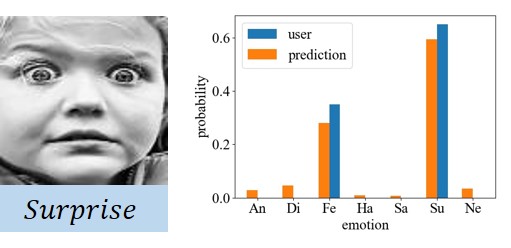}
        \label{00006-vis}
    }
    \subfigure{
        \includegraphics[width=0.3\textwidth]{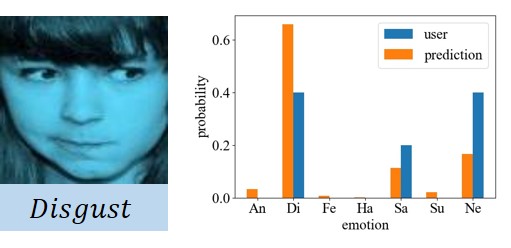}
        \label{04632-vis}
    }
    \subfigure{
        \includegraphics[width=0.3\textwidth]{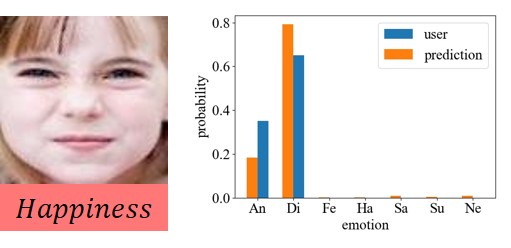}
        \label{05315-vis}
    }
    \caption{Comparison between labels, user study results, and generated label distributions. Please refer to the appendix for more results.}
    \label{fig:target-vis}
\end{figure*}

\begin{figure}[t]
    \centering
    \includegraphics[width=0.4\textwidth]{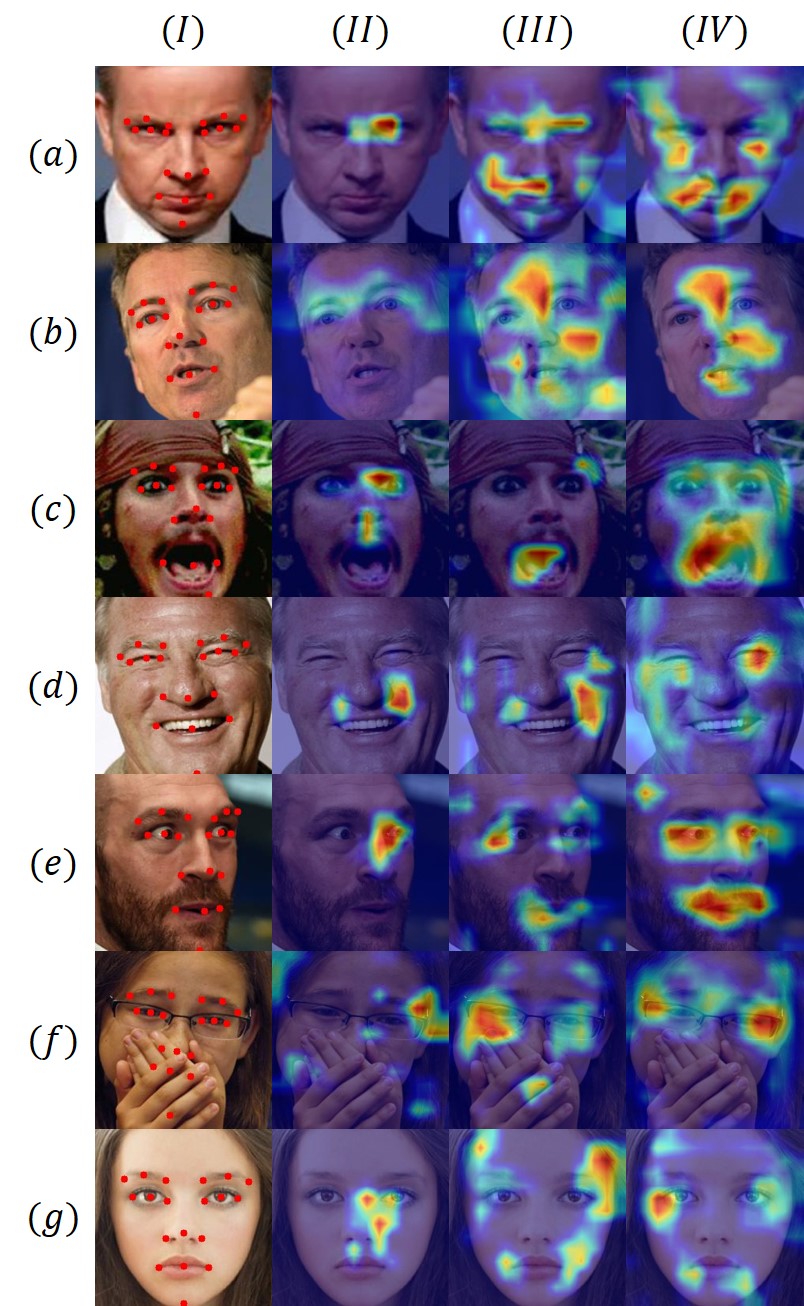}
    \caption{Attention maps of images from AffectNet. Row (a) - (g) denote anger, disgust, fear, happiness, sadness, surprise, and neutral respectively. Column (I) shows the images and the landmarks. Columns (II) - (IV) present the attention maps of three training strategies: (II) Baseline strategy; (III) Using LDE to generate target distribution; (IV) Complete LA-Net.}
    \label{fig:attention-vis}
\end{figure}

\noindent\textbf{Target label distributions.} We conduct a user study on several images randomly selected from RAD\_DB and AffectNet to diagnose the LDE module. We present part of the results in~\cref{fig:target-vis}, and \textit{more details are provided in Appendix C}. The left two expressions in~\cref{fig:target-vis} are found to be ambiguous in the user study, while the one-hot labels only provide one possible category. In contrast, LDE reveals all possible classes and generates label distributions that are consistent with the user study results. Additionally, the model correctly identifies the latent truth for the third image, which is mistakenly assigned to \textit{happiness} in the dataset (highlighted in red). In conclusion, the results indicate that LA-Net can achieve agreement with human perception to some extent and effectively mitigate label noise.

\noindent\textbf{Attention visualization.} To further analyze how landmark information benefits LA-Net, we randomly select some images from AffectNet and generate their attention maps using GradCAM~\cite{selvaraju2017grad}. Note that we perform visualization in the same setting as in inference mode. That is, no landmark information is provided when extracting attention maps. 

\cref{fig:attention-vis} plots the attention maps of the selected images, where the categories are anger, disgust, fear, happiness, sadness, surprise, and neutral from top to bottom. Column (I) provides the images and their landmarks, and the rest presents the attention maps of three strategies. Comparing different columns, the baseline model typically focuses on a few areas, while the LDE module reveals more crucial areas by providing better supervision in the presence of label noise. Moreover, EL Loss contributes to locating more key regions, such as the eyes in (c, IV) and (d, IV), and boosts the attention of key areas, including the eyes and mouth in (e, IV). Overall, LA-Net generates regions of interest that align well with facial landmarks, indicating that it develops a knowledgeable feature extractor by incorporating landmark information into expression representations.

\section{Conclusion}
\label{sec:conclusion}
This paper introduces a new FER model named LA-Net, which aims to mitigate the impact of label noise by incorporating landmark information. LA-Net consists of two main modules, namely label distribution estimation and expression-landmark contrastive loss. The former uses landmark information to correct errors in expression space and estimates the label distribution of each sample, which in turn provides better supervision. The latter enables noise-tolerant supervised contrastive learning and develops a more robust feature extractor by performing interactions between expressions and landmarks. We conduct extensive experiments on multiple scenarios, whose results demonstrate the effectiveness of LA-Net in handling label noise and inconsistent labels. Additionally, we present various visualization studies to illustrate how landmark information enhances the performance of LA-Net.

{\small
\bibliographystyle{ieee_fullname}
\bibliography{egbib}
}

\newpage
\appendix
\begin{table}[t]
    \scriptsize
    \centering
    \begin{tabular}{c|c|c c c}
    \hline
    \textbf{Asymmetric Noise} & Method & RAF-DB & FERPlus & AffectNet \\
    \hline
    \multirow{5}*{10\%} & Baseline & 80.93{\tiny±0.49} & 82.98{\tiny±0.09} & 57.05{\tiny±0.26}\\
    ~ & SCN & 81.55{\tiny±0.28} & 83.02{\tiny±0.16} & 58.49{\tiny±0.33} \\
    ~ & RUL & 85.53{\tiny±0.32} & 85.12{\tiny±0.29} & 60.09{\tiny±0.10} \\
    ~ & EAC & 87.44{\tiny±0.18} & 86.11{\tiny±0.14} & 61.03{\tiny±0.15} \\
    ~ & LA-Net & \textbf{88.53{\tiny±0.20}} & \textbf{87.21{\tiny±0.12}} & \textbf{62.45{\tiny±0.17}}\\
    \hline
    \multirow{5}*{20\%} & Baseline & 75.62{\tiny±0.39} & 80.97{\tiny±0.27} & 55.91{\tiny±0.19} \\
    ~ & SCN & 79.53{\tiny±0.42} & 82.03{\tiny±0.09} & 57.03{\tiny±0.09}\\
    ~ & RUL & 83.55{\tiny±0.29} & 83.76{\tiny±0.14} & 58.45{\tiny±0.21}\\
    ~ & EAC & 85.09{\tiny±0.31} & 84.61{\tiny±0.28} & 59.85{\tiny±0.29}\\
    ~ & LA-Net & \textbf{86.31{\tiny±0.21}} & \textbf{85.44{\tiny±0.17}} & \textbf{61.58{\tiny±0.18}}\\
    \hline
    \multirow{5}*{30\%} & Baseline & 70.38{\tiny±0.62} & 76.09{\tiny±0.29} & 50.84{\tiny±0.30}\\
    ~ & SCN & 74.29{\tiny±0.39} & 80.27{\tiny±0.17} & 54.56{\tiny±0.21}\\
    ~ & RUL & 78.78{\tiny±0.44} & 80.99{\tiny±0.14} & 56.08{\tiny±0.11}\\
    ~ & EAC & 79.62{\tiny±0.18} & 82.09{\tiny±0.11} & 58.50{\tiny±0.19}\\
    ~ & LA-Net & \textbf{81.03{\tiny±0.24}} & \textbf{83.62{\tiny±0.16}} & 
    \textbf{60.19{\tiny±0.22}}\\
    \hline
    \end{tabular}
    \caption{Evaluation~(\%) on asymmetric noise (anger $\rightarrow$ disgust, disgust $\rightarrow$ anger, fear $\rightarrow$ surprise, happiness $\rightarrow$ neutral, sadness $\rightarrow$ neutral, surprise $\rightarrow$ anger, neutral $\rightarrow$ sadness, contempt $\rightarrow$ neutral). We reproduce SCN, RUL, and EAC and report the performance, as they do not consider asymmetric noise in their studies.}
    \label{tab:asymmetric}
\end{table}

\begin{figure}[t]
    \centering
    \subfigure{
       \includegraphics[width=0.22\textwidth]{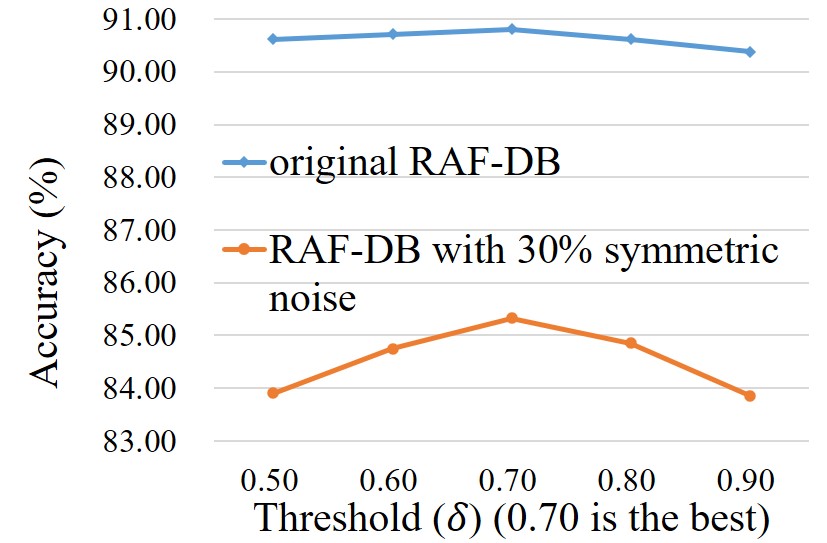}
        \label{threshold}
    }
    \subfigure{
        \includegraphics[width=0.22\textwidth]{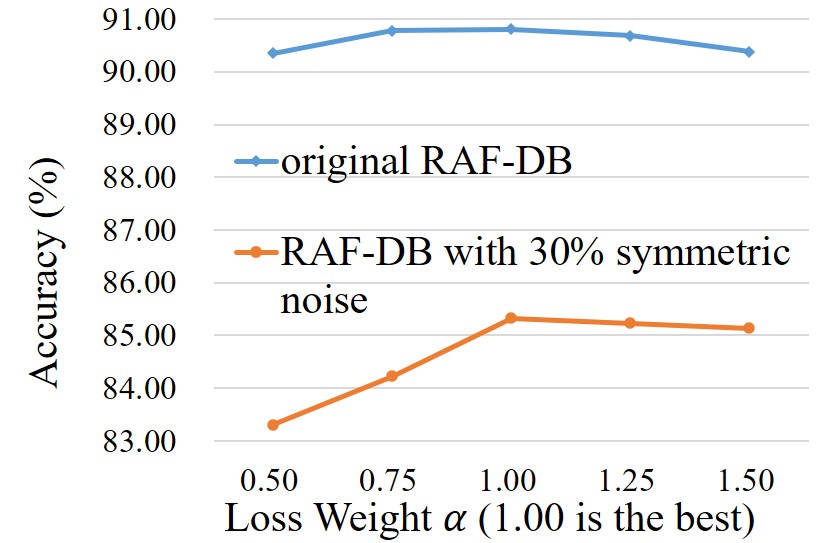}
        \label{alpha}
    }
    \subfigure{
        \includegraphics[width=0.22\textwidth]{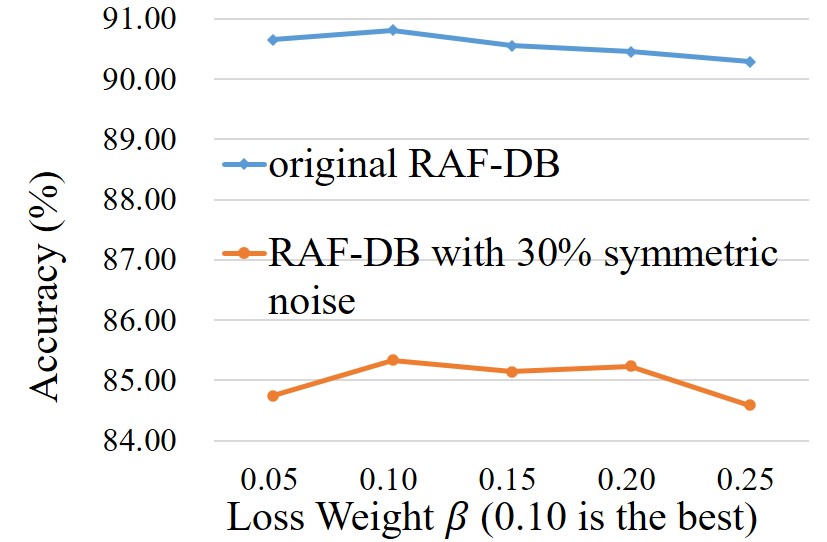}
        \label{beta}
    }
    \caption{Performance with varying parameters and noise levels.}
    \label{fig:parameter}
\end{figure}

\section{Performance with Asymmetric Noise}
\label{sec:asymmetric}
\cref{tab:synthetic-noise} follows prior studies and generates noisy labels by randomly flipping the label to other classes uniformly, referred to as symmetric noise. Given that symmetric noise may not reflect real-world ambiguity, we test asymmetric noise, where the label is flipped to its most similar class based on the confusion matrix. As shown in \cref{tab:asymmetric}, LA-Net consistently outperforms previous models when dealing with asymmetric noise. Specifically, compared to EAC, the approach achieves an average promotion of 1.24\%, 1.15\%, and 1.61\% on the three datasets. Overall, the impressive robustness against asymmetric noise demonstrates the potential of LA-Net to be deployed in real-world applications.

\begin{figure*}[t]
    \centering
    \subfigure{
        \includegraphics[width=0.275\textwidth]{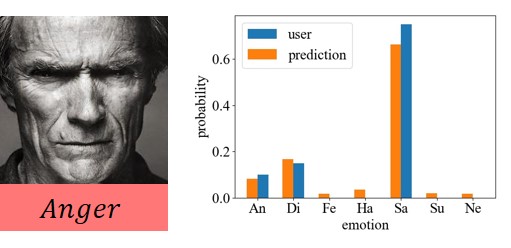}
        \label{7}
    }
    \subfigure{
        \includegraphics[width=0.275\textwidth]{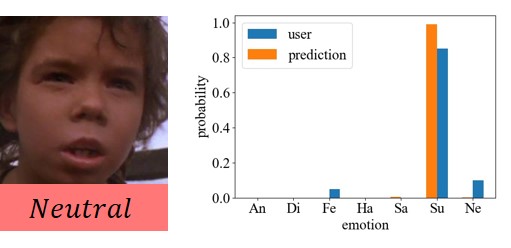}
        \label{2622}
    }
    \subfigure{
        \includegraphics[width=0.275\textwidth]{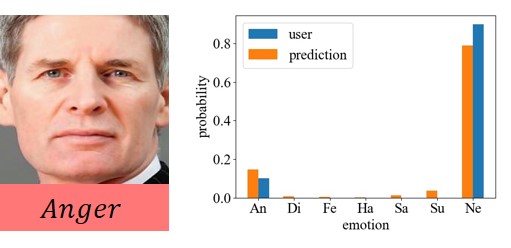}
        \label{2775}
    }
    \subfigure{
        \includegraphics[width=0.275\textwidth]{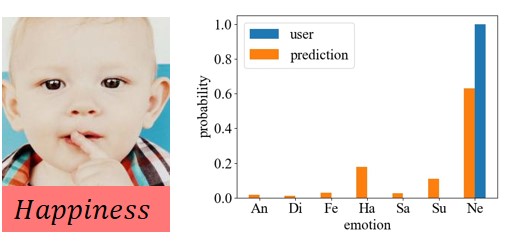}
        \label{3464}
    }
    \subfigure{
        \includegraphics[width=0.275\textwidth]{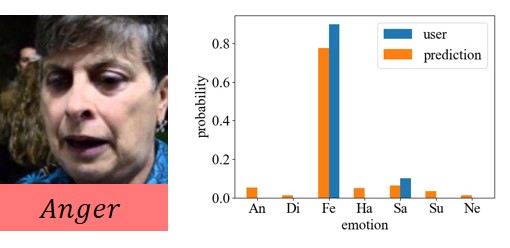}
        \label{475}
    }
    \subfigure{
        \includegraphics[width=0.275\textwidth]{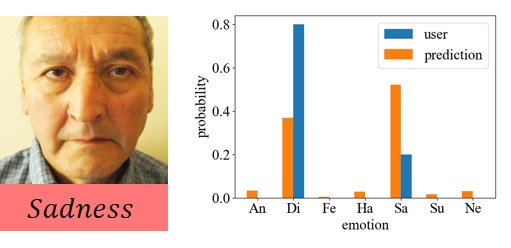}
        \label{43}
    }
    \subfigure{
        \includegraphics[width=0.275\textwidth]{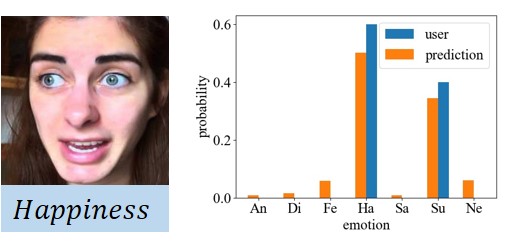}
        \label{155}
    }
    \subfigure{
        \includegraphics[width=0.275\textwidth]{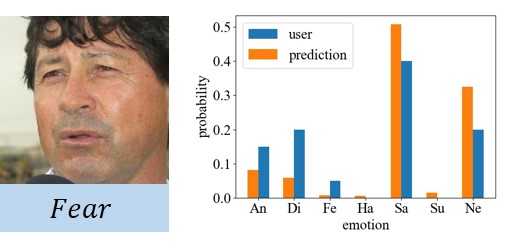}
        \label{585}
    }
    \subfigure{
        \includegraphics[width=0.275\textwidth]{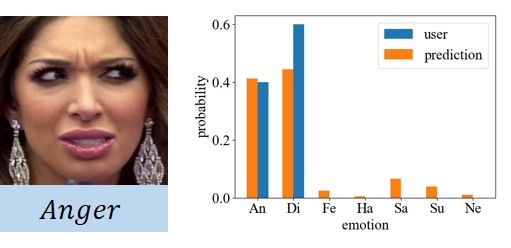}
        \label{912}
    }
    \subfigure{
        \includegraphics[width=0.275\textwidth]{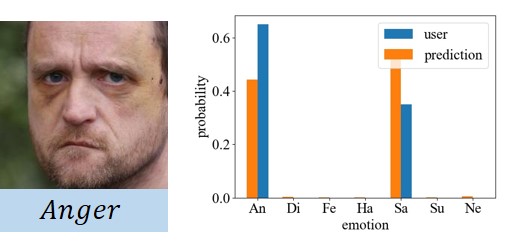}
        \label{1506}
    }
    \subfigure{
        \includegraphics[width=0.275\textwidth]{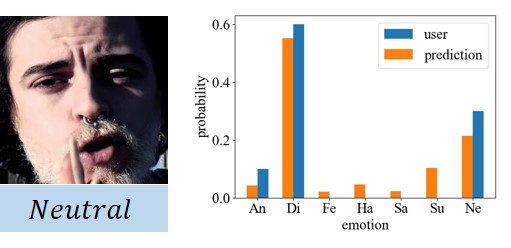}
        \label{5720}
    }
    \subfigure{
        \includegraphics[width=0.275\textwidth]{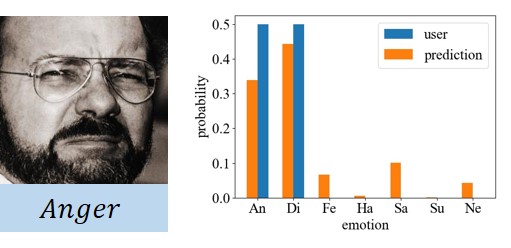}
        \label{6225}
    }
    \subfigure{
        \includegraphics[width=0.275\textwidth]{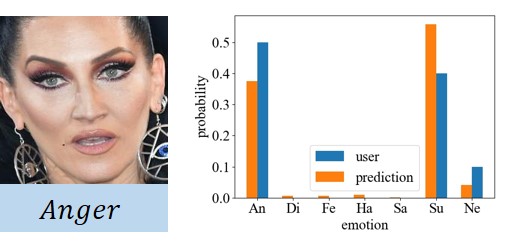}
        \label{5}
    }
    \subfigure{
        \includegraphics[width=0.275\textwidth]{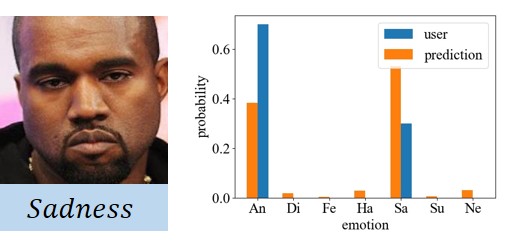}
        \label{42}
    }
    \subfigure{
        \includegraphics[width=0.275\textwidth]{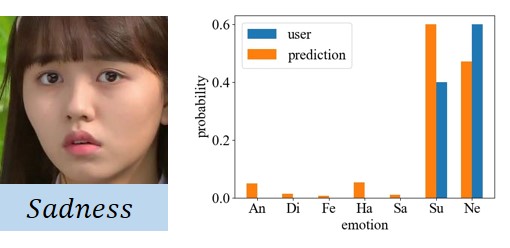}
        \label{7201}
    }
    \subfigure{
        \includegraphics[width=0.275\textwidth]{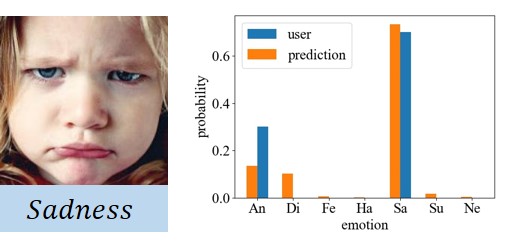}
        \label{7351}
    }
    \subfigure{
        \includegraphics[width=0.275\textwidth]{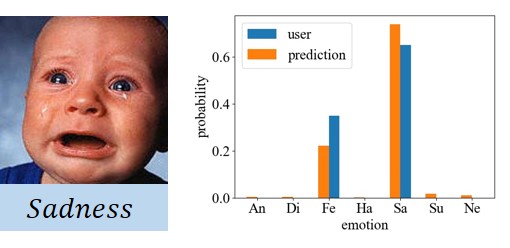}
        \label{7970}
    }
    \subfigure{
        \includegraphics[width=0.275\textwidth]{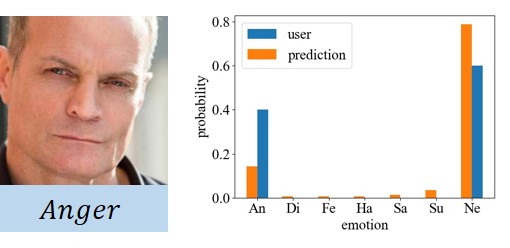}
        \label{8614}
    }
    \subfigure{
        \includegraphics[width=0.275\textwidth]{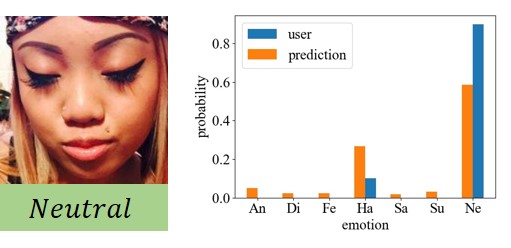}
        \label{2}
    }
    \subfigure{
        \includegraphics[width=0.275\textwidth]{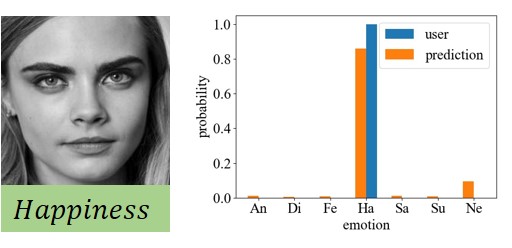}
        \label{10}
    }
    \subfigure{
        \includegraphics[width=0.275\textwidth]{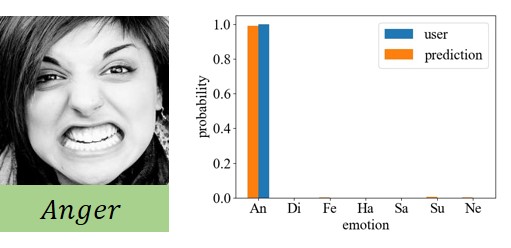}
        \label{18}
    }
    \subfigure{
        \includegraphics[width=0.275\textwidth]{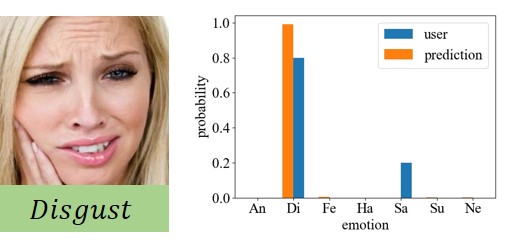}
        \label{4123}
    }
    \subfigure{
        \includegraphics[width=0.275\textwidth]{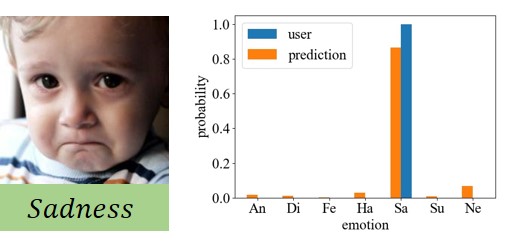}
        \label{4640}
    }
    \subfigure{
        \includegraphics[width=0.275\textwidth]{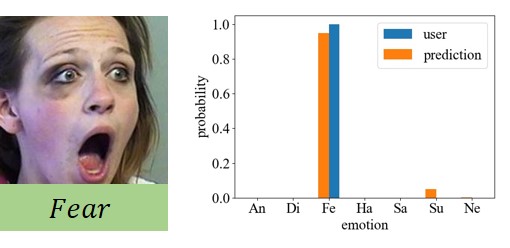}
        \label{4682}
    }
    \caption{More user study results of the images from AffectNet. We highlight the mistakenly annotated images, ambiguous images, and images with correct labels in red, blue, and green, respectively.}
    \label{fig:user}
\end{figure*}

\section{More Grid Search Results.} 
We test various temperature values $\tau$ (0.05, 0.07, 0.1, 0.15, 0.2) and find 0.1 is optimal. We test various momentum decay $\omega$ (0.8, 0.85, 0.9, 0.95, 0.99) and find 0.9 is optimal. We present the grid search for threshold $\delta$ and weights $\alpha$, $\beta$ in \cref{fig:parameter}.

\section{More User Study Results}
We conduct a user study using 50 images selected from AffectNet~\cite{mollahosseini2017affectnet} and compare the one-hot labels, human perception, and generated label distributions to evaluate the effectiveness of our LA-Net. To present our findings more succinctly, we plot all of the mistakenly annotated and ambiguous images, as well as a part of the correctly labeled images (24 in total) in~\cref{fig:user}. 

The first two lines present several images that are mistakenly annotated in AffectNet~(highlighted in red). LA-Net identifies these errors and reveals the latent truth. Moreover, images plotted in rows 3-6 are found to be ambiguous according to user study results~(highlighted in blue). Fortunately, the proposed model generates label distributions consistent with human perception, reducing the uncertainty of these ambiguous samples. Additionally, as shown in the last two lines of~\cref{fig:user}, LA-Net produces targets that align well with the one-hot labels for correctly annotated samples~(highlighted in green). Overall, the model demonstrates some level of agreement with human perception and effectively mitigates label noise.

In addition to visualization, we quantitatively evaluate the consistency between the label distributions and the user study results of the selected images using Jensen-Shannon divergence~(\textit{JS} divergence). Mathematically, given probability distributions $\bm{p_1}$ and $\bm{p_2}$, their \textit{JS} divergence can be calculated by:

{
\footnotesize
\begin{align}
    JS(\bm{p_1}||\bm{p_2}) &= \frac{1}{2}KL(\bm{p_1}||\frac{\bm{p_1}+\bm{p_2}}{2}) + \frac{1}{2}KL(\bm{p_2}||\frac{\bm{p_1}+\bm{p_2}}{2})
\end{align}
}
\cref{tab:js} reveals a significant difference between one-hot labels and user study results,
indicating that FER datasets~\cite{li2017reliable, mollahosseini2017affectnet, barsoum2016training} suffer from serious label noise. In contrast, LA-Net generates label distributions that are in better agreement with human perception. This improvement leads to better FER performance, especially when dealing with label noise.

\begin{table}[h]
    \centering
    \begin{tabular}{c|c}
    \hline
     & user study \\
    \hline
    \hline
    one-hot labels & 0.2030 \\
    label distributions & \textbf{0.0909}\\
    \hline
    \end{tabular}
    \caption{Jensen-Shannon divergence between the user study results and two expression descriptors.}
    \label{tab:js}
\end{table}

\end{document}